%% file: emnlp2020.tex
\pdfoutput=1
\documentclass[11pt,a4paper]{article}
\usepackage[hyperref]{emnlp2020}
\usepackage{times}
\usepackage{latexsym}
\usepackage{booktabs}
\usepackage{graphicx}

\usepackage{times}
\usepackage{latexsym}
\usepackage{url}
\usepackage{textcomp}
\usepackage{bbm}
\usepackage{amsmath}
\usepackage{booktabs}
\usepackage{tabularx}
\usepackage{graphicx}
\usepackage{dialogue}
\usepackage{mathtools}
\usepackage{hyperref}
\usepackage{comment}

\usepackage{multirow}
\usepackage{mdframed}
\usepackage{tcolorbox}
\usepackage{enumitem}

\usepackage{xcolor,pifont}

\newcommand\cmark {\textcolor{green}{\ding{52}}}
\newcommand\xmark {\textcolor{darkgray}{\ding{55}}}
\newcommand\omark {\textcolor{brown}{\ding{60}}}

\usepackage{amssymb}
\usepackage{soul}
\makeatletter
\def\blfootnote{\xdef\@thefnmark{}\@footnotetext}
\makeatother
\usepackage{microtype}

\aclfinalcopy 

\title{Benchmarking Robustness of Machine Reading Comprehension Models}

\author{Chenglei Si$^{1*}$ , Ziqing Yang$^{2}$, Yiming Cui$^{2,3}$, Wentao Ma$^{2}$, Ting Liu$^{3}$, Shijin Wang$^{2,4}$ \\
{$^{1}$ University of Maryland, College Park, MD, USA} \\
{$^{2}$ Joint Laboratory of HIT and iFLYTEK Research (HFL), Beijing, China}\\
{$^{3}$ Research Center for SCIR, Harbin Institute of Technology, Harbin, China}\\
{$^{4}$ State Key Laboratory of Cognitive Intelligence, Beijing, China}\\
{$^{1}$\tt clsi@terpmail.umd.edu} \\
{$^2$\tt\{zqyang5,ymcui,wtma,sjwang3\}@iflytek.com}\\  
{$^3$\tt \{ymcui,tliu\}@ir.hit.edu.cn}\\
}

\date{}

\begin{document}
\maketitle
\begin{abstract}
Machine Reading Comprehension (MRC) is an important testbed for evaluating models' natural language understanding (NLU) ability. There has been rapid progress in this area, with new models achieving impressive performance on various benchmarks. However, existing benchmarks only evaluate models on in-domain test sets without considering their robustness under test-time perturbations or adversarial attacks. To fill this important gap, we construct AdvRACE (Adversarial RACE), a new model-agnostic benchmark for evaluating the robustness of MRC models under four different types of adversarial attacks, including our novel distractor extraction and generation attacks. We show that state-of-the-art (SOTA) models are vulnerable to all of these attacks. We conclude that there is substantial room for building more robust MRC models and our benchmark can help motivate and measure progress in this area.
We release our data and code at \url{https://github.com/NoviScl/AdvRACE}.
\end{abstract}

{\let\thefootnote\relax\footnotetext{$^*$ Work done during an internship at HFL.}}

\begin{table*}[t]
 \small
 \centering
\begin{tabular}
{l|c|c|c|c|c|c}
\toprule
Method & AddSent & CharSwap & WordReplace & Paraphrase & DE & DG  \\
\midrule
\textbf{AdvRACE (Ours)} $^{\dagger}$ & \cmark & \cmark & \omark & \omark & \cmark & \cmark \\
\midrule
AddSent~\cite{AdvSquad} $^{\dagger}$ & \cmark & \xmark & \xmark & \xmark & \xmark & \xmark \\
SEA~\cite{SEA} $^{\dagger}$ & \xmark & \xmark & \xmark & \cmark & \xmark & \xmark \\
SCPN~\cite{SCPN} $^{\dagger}$ & \xmark & \xmark & \xmark & \cmark & \xmark & \xmark \\
charNMT~\cite{charNMT} $^{\dagger}$ & \xmark & \cmark & \xmark & \xmark & \xmark & \xmark  \\
breakNLI~\cite{breakNLI} $^{\dagger}$ & \xmark & \xmark & \cmark & \xmark & \xmark & \xmark \\
HotFlip~\cite{Hotflip} & \xmark & \cmark & \cmark & \xmark & \xmark & \xmark \\
Genetic~\cite{genAdv} & \xmark & \xmark & \cmark & \xmark & \xmark & \xmark  \\
PWWS~\cite{PWWS} & \xmark & \xmark & \cmark & \xmark & \xmark & \xmark  \\
UniversalTriggers~\cite{uniTrigger} & \cmark & \xmark & \xmark & \xmark & \xmark & \xmark  \\
ORB~\cite{ORB} $^{\dagger}$ & \xmark & \xmark & \cmark & \cmark & \xmark & \xmark \\
TextFooler~\cite{TextFooler} & \xmark & \xmark & \cmark & \xmark & \xmark & \xmark\\
SememePSO~\cite{combiOpt} & \xmark & \xmark & \cmark & \xmark & \xmark & \xmark \\
\bottomrule 
\end{tabular}
\caption{Comparison of AdvRACE to other adversarial attack work. {\omark} indicates that the method was tried but not included in the benchmark due to poor performance. $^{\dagger}$ indicates these methods are model-agnostic.} 
\vspace{-10pt}
 \label{tab:dataset_comparison}
\end{table*}

\section{Introduction}
The goal of Machine Reading Comprehension (MRC) is to examine whether the model can understand the text and perform certain types of reasoning. 
To this end, many MRC benchmarks have been constructed in different domains, styles, and languages~\cite[][\textit{inter alia}]{Squad, Race, DROP, CMRC}.
While most of these benchmarks use leaderboards to compare different models' performance on in-domain test sets, they have ignored the important aspect of evaluating models' robustness. 

The research on robustness of MRC models can be generally categorised into two directions: generalization to out-of-domain distributions and robustness under test-time perturbations.
Regarding generalization to out-of-domain distributions, \citet{multiqa} and the MRQA shared task~\cite{mrqa} investigated how well do MRC models trained on source datasets generalize on unseen datasets in different domains.

On evaluating MRC models under test time perturbations, 
previous works~\cite{AdvSquad, paraQ} evaluate models under only one specific attack. However, truly robust models should perform well on different types of perturbed test inputs instead of just one. 
In this work, we cover diverse types of adversarial attacks to allow for a more comprehensive evaluation of model robustness.
\citet{ORB} performed a set of synthetic perturbations to test models' robustness. However, their perturbations are only applicable to specific types of questions, yielding limited number of valid examples, limiting their usefulness in revealing model weakness. This may be because that they only performed perturbations on the questions. In contrast, our proposed methods are applied to different components (passages, questions, distractors) of the dataset and at different levels (sentence and character levels). In section~\ref{sec:model_eval}, we show that all of our perturbations incur significant performance drops on SOTA models, leaving large room for future improvement. 


Instead of constructing a new dataset from scratch, we leverage on an existing MRC dataset RACE~\cite{Race}, where we apply our proposed attacks on the RACE test set to form our AdvRACE benchmark.
We choose to construct our benchmark based on RACE because: 1) The multiple-choice format supports more types of attacks. We generated new distractors to replace the original ones as a novel attack method. 2) RACE covers a diverse set of linguistic phenomena and reasoning types, and has been used widely for evaluating NLU performance~\cite{XLNet,Roberta,ALBERT}. 

 Our AdvRACE benchmark contains four test sets for four different adversarial attacks: AddSent, CharSwap, Distractor Extraction (DE) and Distractor Generation (DG). 
 The performance under each attack allows for fine-grained and comprehensive analysis of model robustness.
In addition, the advantages of our work include: 
1) The pipeline we use to construct AdvRACE is highly efficient and transferable.
It can be easily adopted on other MRC datasets to construct the corresponding adversarial test sets. 
2) All of our adversarial perturbations are model-agnostic and do not require access to model parameters. Thus it allows for the evaluation of any model to reveal their weaknesses in robustness. 
3) Our perturbations are label-preserving. Our human annotations ensured the high quality of our adversarial test sets (section \ref{human_eval}).

\section{Related Work}
\textbf{Adversarial attacks in NLP.}
Various adversarial attack methods have been proposed for NLP tasks. One type is white-box attack where perturbations are constructed based on models' gradients or parameters to exploit their weakness~\cite{genAdv, Hotflip, uniTrigger}. However, such attacks cannot be directly applied when there is no access to model internal parameters. Another type is black-box attack where the perturbation is constructed without accessing model parameters and is often based on heuristic rules or produced by specifically designed models~\cite{AdvSquad, SEA, SCPN}. While the above approaches construct attacks automatically, there are also attempts on human-in-the-loop adversarial example generation where human annotators are employed to write new test data that can fool the models into making wrong predictions~\cite{ANLI, trickme}. However, such approach requires significant human and time resources, which may not be easily accessible.


\noindent
\textbf{Robustness benchmarks for MRC.}
\citet{DuReaderRobust} created a Chinese MRC robustness benchmark by constructing distracting questions via retrieval from existing database and human writing. \citet{benchmarkSquad} applied various common synthetic perturbations on the SQuAD dataset to analyse the factors affecting model robustness. \citet{CheckList} provided a general framework for automatically identifying various types of failures in NLP systems with CheckList. Our adversarial attacks have similar goals as the invariance test in CheckList. However, their template-based checks only examine limited failure types, thus less effective in revealing weaknesses in model robustness as shown by the relative small drop of model performances under invariance tests, as compared to our proposed  attacks on MRC.



\begin{table*}[t]
\centering \footnotesize
\begin{tabular}{p{0.005\textwidth}p{0.93\textwidth}}

 \toprule
 \parbox[t]{1mm}{\multirow{2}{*}{\rotatebox[origin=c]{90}{{\textbf{Original}}}}} &
\textbf{Passage:} 
        ``Thanks for coming," Everett said, shaking hands with Mr. Hanson, the town councilor.  "I'm curious about the ideas in your letter."  Mr. Hanson nodded toward the parking lot near where they stood. ...
        Everett was relieved that the community representative seemed receptive to the idea. Mr. Hanson studied the photograph, and then asked, "If town council provides the money, how will you and your friends contribute?" Everett felt optimistic now.
        \\ & 
        \textbf{Question:} Everett was trying to persuade Mr. Hanson to  \_  . \\  & 
\textbf{Answer:} provide money for a skateboard park \\
& \textbf{Options:}  
\textbf{1.} beautify the neighbourhood 
 \textbf{2.} build an art centre for children
 \textbf{3.} cut the area of the parking lot  \\

\midrule
\parbox[t]{1mm}{\multirow{2}{*}{\rotatebox[origin=c]{90}{{\textbf{AddSent}}}}} &
\textbf{Passage:} 
``Thanks for coming," Everett said, shaking hands with Mr. Hanson, the town councilor. "I'm curious about the ideas in your letter." Mr. Hanson nodded toward the parking lot near where they stood. "Please tell me more." Everett took a deep breath. "Ever since the school closed two years ago, this area has become worse. But if we clean up the litter and repair the fence, it will be a great place for a skateboard park." \textbf{\textit{Zimmerman was trying to dissuade Mr. Morrison to build an art centre for children.}} Mr. Hanson scanned the broken concrete, nodding. ...
Everett was relieved that the community representative seemed receptive to the idea. \textbf{\textit{Larkin was trying to dissuade Mr. Lyons to beautify the neighbourhood.}} Mr. Hanson studied the photograph, and then asked, "If town council provides the money, how will you and your friends contribute?" Everett felt optimistic now.
\\ & \textbf{Question/Answer/Options:} Same as \textbf{Original}. \\

\midrule
\parbox[t]{1mm}{\multirow{2}{*}{\rotatebox[origin=c]{90}{{\textbf{CharSwap}}}}} & \textbf{Passage:} ``Thanks for coming," Everett said, shaking hands with Mr. Hasnon, the town \textbf{\textit{councilor}}.``I'm curious about the ideas in your letter." Mr. Hanson nodded toward the \textbf{\textit{pakring}} lot near where they stood. ``Please tell me more. "Everett took a deep breath.``Ever since the school closed two years ago, this \textbf{\textit{aera}} has become worse . But if we clean up the litter and repair the fence, it will be a great place for a \textbf{\textit{skateborad}} park." Mr. Hanson scanned the broken concrete, nodding. ``The old school is being adapted to a community arts \textbf{\textit{cenrte}}. This \textbf{\textit{aera}} could become a vital part of the \textbf{\textit{neighbouhrood}} again." ...
\\
&\textbf{Question:} Everett was \textbf{\textit{tyring}} to \textbf{\textit{perusade}} Mr. Hanson to  \_  .
\\
&\textbf{Answer/Options:} Same as \textbf{Original}. \\

\midrule
\parbox[t]{1mm}{\multirow{2}{*}{\rotatebox[origin=c]{90}{{\textbf{DE}}}}} &
\textbf{Passage/Question/Answer:} Same as \textbf{Original}. \\
& \textbf{Options:} \textbf{1.} build the ramps
 \textbf{2.} provide 
 \textbf{3.} repair the fence, it will be a great place for a skateboard park \\

\midrule
\parbox[t]{1mm}{\multirow{2}{*}{\rotatebox[origin=c]{90}{{\textbf{DG}}}}} &
\textbf{Passage/Question/Answer:} Same as \textbf{Original}. \\
& \textbf{Options:} 
 \textbf{1.} help the town councilor 
 \textbf{2.} help design and build the ramps
 \textbf{3.} buy the old school \\
\bottomrule
\end{tabular}
\caption{Examples of each attack applied on the same original test example. \textbf{\textit{Italic}} parts are altered by our adversarial attack perturbations.}
\label{tab:examples}
\end{table*}

\section{Adversarial Attacks}
\label{methods}
In this section, we describe the different types of adversarial attacks explored in AdvRACE. 
In Table~\ref{tab:dataset_comparison}, we present a comparison of our AdvRACE benchmark with other works on automatic construction of adversarial attacks.

In multiple-choice MRC, distractors naturally play an important part in affecting the difficulty of the dataset. Hence, we propose a novel type of adversarial attack method by constructing new sets of distractors to replace the original ones. We present two different methods of constructing new distractors: distractor extraction (DE) and distractor generation (DG).
Apart from them, we have also experimented four other types of adversarial attack methods: distracting information insertion (AddSent), character swap (CharSwap), word replacement and paraphrase. 
Among them, we find that WordReplace and Paraphrase do not work well on RACE, hence being excluded from our final AdvRACE benchmark.
We provide examples of each adversarial attack in Table~\ref{tab:examples}. The details of the two failed attempts can be found in the Appendix.

\begin{table*}[h]
\centering
\small 
\begin{tabular}{ lccccc|cc }
\toprule
 & \textbf{BERT}
 & \textbf{RoBERTa}
 & \textbf{XLNet} 
 & \textbf{ALBERT}
 & \textbf{\textit{Average }}
 & \textbf{Valid}
 & \textbf{Correct}
 \\
 \midrule
Original & 68.5 & 83.7 & 79.9 & 86.0 & & 100.0\% & 100.0\% \\
AddSent & 25.5 \small{(\textit{-62.8\%})} & 62.3 \small{(\textit{-25.6\%})} & 54.1 \small{(\textit{-32.3\%})}& 69.2 \small{(\textit{-19.5\%})} & \small{\textit{-35.1\%}} & 98.0\% & 89.8\% \\
CharSwap & 48.8 \small{(\textit{-28.8\%})} & 69.4 \small{(\textit{-17.1\%})} & 63.4 \small{(\textit{-20.7\%})} & 73.0 \small{(\textit{-15.1\%})} & \small{\textit{-20.4\%}} & 100.0\% & 94.0\% \\
Distractor Extraction & 32.9 \small{(\textit{-52.0\%})} & 48.8 \small{(\textit{-41.7\%})} & 44.0 \small{(\textit{-44.9\%})}& 52.0 \small{(\textit{-39.5\%})} & \small{\textit{-44.5\%}} & 98.0\%  & 95.9\% \\
Distractor Generation & 56.3 \small{(\textit{-17.8\%})} & 68.9 \small{(\textit{-17.7\%})}& 64.7 \small{(\textit{-19.0\%})} & 70.9 \small{(\textit{-17.6\%})} & \small{\textit{-18.0\%}} & 98.0\% & 93.9\%  \\
\midrule
Average & 40.9 \small{(\textit{-40.3\%})} & 62.4 \small{(\textit{-25.4\%})} & 56.6 \small{(\textit{-29.2\%})} & 66.3 \small{(\textit{-22.9\%})}  \\
\bottomrule
\end{tabular}
\caption{Left: Attack results on different models. Right: Human evaluation results. \textit{Numbers} in brackets are the percentage drop in performance. Average is computed over all the adversarial test sets for each model, and over all models for each adversarial attack.}
\label{results}
\end{table*}






\subsection{Distractor Extraction}
For distractor extraction, we aim to extract spans from the passages as new distractors. This is motivated by the observation that most passages contain distracting sequences that are relevant to the question while semantically different from the correct answer. Such distracting sequences can be especially effective as current MRC models reply largely on text matching~\cite{Si2019WhatDB}.

The key challenge is how do we extract such distracting sequences from the passages without annotated supervised data. In order to solve this challenge, we propose a novel distractor extraction method. We first turn RACE into span-extraction format and adopt the span-extraction QA model used in \citet{BERT} except that we use ALBERT as our backbone model. During training, we insert the correct answers of RACE questions into the passages as the gold answer span. The ALBERT span extractor is trained to select the answer span from the passage. During inference, we use the trained model to extract spans from the passages for the test questions but without inserting the correct answer. In this way, since the answers in RACE have not appeared in the passages in verbatim, there are no gold answer spans for the test examples. However, the extracted spans with high probability can be considered as likely options for the question, which can serve as strong distractors. For post-processing, we select 3 distractors among the top 20 candidate spans that have low lexical overlap with each other and also low lexical overlap with the correct answer, so that they are diverse and label-preserving.



\subsection{Distractor Generation}
Another way to construct new distractors is to generate them based on the passage and question. We frame it as a sequence-to-sequence (Seq2Seq) task where the input is the concatenation of the passage and question, and the output is the distractor. Previous distractor generation works \cite{DisGen,DGsota} adopted this approach. However, none of these works has explored using the generated distractors as a way for attacking MRC models. In our work, we use the filtered data from \citet{DisGen} for training the DG model, where distractors with low semantic relevance with the passages are pruned. 
For the backbone model, we adopt UniLM~\cite{unilm}, a competitive pretrained model unifying NLU and NLG.
We use beam search to find the top $k$ (beam size, we used $k=50$) candidate distractors and select the top 3 among them so that their Jaccard distance between each other is larger than 0.5, to ensure diversity of the 3 selected distractors.
We provide additional DE and DG examples in Table~\ref{tab:DE_DG_examples}.

\subsection{AddSent}
Inspired by the method proposed by \citet{AdvSquad}, we propose an improved AddSent method to make use of the human-written distractors in RACE rather than generating new fake answers to construct strong distracting sequences to insert into the original passages. 
The motivation is to add distracting information that appears similar to the question and the distractors so that it can better mislead models. We use the following procedure to construct the perturbations:
\begin{enumerate}[leftmargin=*] \setlength{\itemsep}{0.001pt}
    \item We change all nouns and numbers in the questions to their nearest word in GloVe~\cite{Glove} embedding space with the same part-of-speech. 
    \item We replace adjectives, adverbs, verbs in the questions with their antonym in WordNet~\cite{WordNet} with the same part of speech. We only change one word in each question to its antonym to prevent confusions caused by multiple negations (e.g., `likely to obey' being changed to `unlikely to disobey').
    \item If no words are changed during Step 1 and 2 (28.9\% of test questions), we insert the negation word `not' at the appropriate position in the question. 
    The first three steps ensure that there is a flip in the semantic meaning of the original question.
    \item We randomly sample a distractor from the original three distractors of the question, and concatenate it with the altered question.
    \item We insert the concatenated sequence into a random position of the passage.
    \item We repeat Step 1 to 5 one more time with different replacement words from GloVe and WordNet, and using a different distractor. This makes the perturbation more misleading for the model since there is more matching.
\end{enumerate}



\subsection{CharSwap}
It is shown in \citet{charNMT} that NMT performance drops significantly when there are spelling errors in the data. We adapt their approach in MRC to swap two adjacent letters in a word without altering the first or last letters. Although they show that with more tokens altered, the performance gets worse, empirically we find that applying such perturbation to all words in the dataset sometimes impacts the readability of the text and cause difficulty even for humans to perform well. As a result, we only apply the CharSwap perturbation to the following words with character length of four and above: 1) The non-stopwords in the question. 2) Non-stopwords in the passage that have also appeared in the question and its corresponding options. This is based on the observation that these words are usually keywords for solving the question. There are a total of 7.1\% words being altered using this method.

\section{Evaluation}
We apply the four attacks in Section~\ref{methods} on the original RACE test set to form AdvRACE, which results in four sets of adversarial test sets: AddSent, CharSwap, DE, and DG. 
Each adversarial test set has 4,934 examples, same as the original RACE test set.
We perform human and model evaluations on these test sets. 

\subsection{Human Evaluation}
\label{human_eval}
We randomly sample 50 questions from each adversarial test set in AdvRACE as well as the original test set, and ask paid corporate professional data annotators to answer the MRC questions. To assess the validity of our adversarial data, we add an additional option for each question: unanswerable. If the annotator chooses unanswerable for a question, we further ask the annotator to choose the specific reason why the question is unanswerable, which includes the following options: missing key information, conflicting information, un-interpretable content, no correct options, multiple correct options.
The results are shown in the last 2 columns of Table~\ref{results}. \textbf{Valid} rate measures how many examples are considered by the annotators as interpretable (semantically and syntactically acceptable) and answerable. \textbf{Correct} rate measures how many examples are answered correctly by the annotators.
All the four adversarial test sets have high valid rate close to 100\%. The correct rate of the four adversarial test sets are also above the turker performance of 73.3\% as reported in \citet{Race}. These indicate that the four adversarial test sets in AdvRACE remain well solvable for humans.
We also manually analysed the examples answered wrongly by the human annotators. For all the four adversarial test sets, we find that the drop in human performance (correct rate) is due to unrelated human errors (e.g., misunderstanding of the original text) rather than caused by our perturbations.

\subsection{Model Evaluation}
\label{sec:model_eval}
 For model evaluation, all the models are trained on the original RACE training set, tuned hyper-parameters on the original RACE dev set, and tested on our AdvRACE benchmark.
 We evaluate four competitive models: BERT-Large, XLNet, RoBERTa-Large, ALBERT-xxLarge. We concatenate each candidate answer with the corresponding passage and question, then encode each of these four concatenated sequences and pass the representations of the first position of each sequence (i.e., representations of the \texttt{[CLS]} token) through a fully-connected layer for answer prediction.
 The results are presented in Table~\ref{results}.  In addition to reporting the accuracy of each model on each test set, we also report the percentage performance drop relative to the performance on the original test set for the adversarial test sets. 

From the results, we find that: 
1) BERT is the most vulnerable among these four models (by comparing the average percentage drop in performance in the last row). Models that achieve higher performance on the original test set generally suffer less relative performance drop from the adversarial attacks. 2) Distractor extraction is more effective than distractor generation in terms of attacking the models, probably due to the fact that these models rely heavily on text matching~\cite{Si2019WhatDB}.
3) All of the adversarial attacks in our AdvRACE benchmark incur a significant drop of models' performance, which is much more effective than previous attempts~\cite{ORB} in revealing weaknesses of the most competitive MRC models.

\subsection{Adversarial Training}

A common baseline to defend such adversarial attacks is to add adversarial examples in training, \textit{i.e.}, adversarial training. We experiment such strategy as a baseline, and find that adversarial training is not a truly reliable defense method as it largely exploits dataset artifacts. We present the detailed adversarial training experiments as well as the follow-up analysis in the Appendix (section~\ref{sec:at}).
We believe that better defense techniques are needed to reliably improve model robustness for MRC.

\section{Conclusion}
In this work, we present AdvRACE, a benchmark that evaluates MRC models under diverse types of adversarial attacks. AdvRACE consists of our proposed distractor extraction and generation attacks as well as improved versions of existing attacks adapted to work effectively on MRC. The performance of four competitive MRC models drops significantly on AdvRACE, indicating the effectiveness of our methods for analysing model weaknesses. We hope our work can motivate the development of more robust models that perform well on different perturbations.


\section*{Impact Statement}
Our AdvRACE dataset was constructed based on the original RACE dataset, which was released under an MIT license, allowing for free reuse. We have also obtained the permission from the original authors of RACE to use it to construct AdvRACE for research purposes. 

\noindent The human annotators involved in this project are all full-time employed corporate annotators, who are well-compensated and work in good working conditions.

\noindent We believe that building robust and reliable MRC models is crucial for them to be deployed in real-life applications. Our work is far from enough to thoroughly measure the true robustness of MRC models. We hope that it will be a first step leading to more efforts along this direction.


\bibliography{emnlp2020}
\bibliographystyle{acl_natbib}

\appendix

\clearpage 
\section*{Appendix}
\label{sec:appendix}

\input{appendix}

\end{document}

%% file: appendix.tex
\section{Failed Attacks}
We introduce the two attacks that we explored but did not work well: Word Replacement and Paraphrase.

\subsection{Word Replacement}
For word replacement attack (WordReplace), we adopt the state-of-the-art method: SememePSO~\cite{combiOpt}, and we used RoBERTa as the target model to generate the adversarial examples. Then we transferred these adversarial examples to attack other models. 
The original SememePSO attack was applied on sentence classification tasks. Since the input of RACE is much longer, in order to keep the running time within budget constraints, we have set the population size to 10 and number of maximum optimization iterations to 5. We have also tried simple  random replacement strategies (randomly select a candidate to replace the original word) which is not effective in attacking the models.
The results are shown in Table~\ref{WordReplace}, we find that although the attack is effective in attacking the target model RoBERTa, it is much less effective on other models, which does not satisfy our model-agnostic requirement, and hence this attack is not included in AdvRACE.

\subsection{Paraphrase}
For Paraphrase, we adopted the Syntactically Controlled Paraphrase Network (SCPN)~\cite{SCPN} which generates paraphrases of a sentence based on the given parse templates. The original method was proposed to attack single-sentence text classification problems.
We generate paraphrases for sentences in the passages as an attack.
We generated the Paraphrase test set with SCPN, using the top two levels of the linearized parse tree of the sentence as the input parse template. We have tried using other templates (such as using the most frequent parse template in the test set passage sentences) that are different from the original parse of the sentences, which results in more syntactic variances, but also lower valid rate.
As shown in Table~\ref{WordReplace}, although it is model-agnostic, during human evaluation, we find that 15 out of the 50 randomly sampled examples are labeled as unanswerable. Among the 15 unanswerable examples, 13 examples are labeled as missing key information, the other 2 are labeled as uninterpretable. This suggests that it is difficult to generate high-quality label-preserving adversarial examples by paraphrasing for MRC. Due to the low validity rate of the paraphrased examples, we exclude Paraphrase in AdvRACE to ensure that our benchmark is solvable for human. 

\begin{table}[h]
\centering
\small 
\setlength{\tabcolsep}{1mm}{
\begin{tabular}{ lcccc }
\toprule
 & \textbf{BERT}
 & \textbf{RoBERTa}
  & \textbf{XLNet}
  &  \textbf{ALBERT}
 \\
 \midrule
Original & 68.5 & 83.7 & 79.9  & 86.0 \\
WordReplace & 64.2 & 70.7 & 74.7 & 80.5 \\
Paraphrase & 59.4 & 72.3 & 68.2 & 73.7 \\
\bottomrule
\end{tabular}}
\caption{Performance on WordReplace and Paraphrase test set in comparison with performance on the original RACE test set.}
\label{WordReplace}
\end{table}


\section{Dataset Statistics}
Each adversarial test set in AdvRACE contains 4934 questions, which corresponds to 1407 passages. The average length of each component is shown in Table~\ref{tab:stats}.

\begin{table}[ht]
    \small 
    \centering
    \begin{tabular}{lccc}
    \toprule
    \textbf{Test Set}
  &  \textbf{Passage Len} 
  &  \textbf{Question Len}
  & \textbf{Option Len} \\
  \midrule
  Original & 316.2 & 11.0 & 6.3 \\
  AddSent & 350.6 & 11.0 & 6.3 \\ 
  CharSwap & 316.2 & 11.0 & 6.3 \\ 
  DE & 316.2 & 11.0 & 6.8 \\ 
  DG & 316.2 & 11.0 & 5.7 \\
  \bottomrule
    \end{tabular}
    \caption{Dataset Statistics of the original RACE test set and the adversarial test sets in AdvRACE.}
    \label{tab:stats}
\end{table}

\section{Model Benchmarking Hyper-parameters}
The four models being benchmarked in the paper are: BERT-large, XLNet, ALBERT-xxlarge and RoBERTa-large. For both finetuning on the original RACE training set and adversarial training, we use learning rate 1.5e-5, weight decay 0.1, warmup ratio 0.06, max sequence length 512, batch size 32, max number of epochs 4. All experiments are done on Tesla V100 GPUs.

\begin{table*}[h]
\centering
\small 
\setlength{\tabcolsep}{3mm}{
\begin{tabular}{ lcccccc }
\toprule
 & \textbf{Original}
 & \textbf{+AddSent}
 & \textbf{+CharSwap}
 & \textbf{+DE}
 & \textbf{+DG}
 & \textbf{Mix}
 \\
 \midrule
Original & 83.7 & 82.1 & 81.5 & 80.0 & 81.1 & 81.0 \\
AddSent & 62.3 & 88.0 & 57.8 & 76.2 & 65.6 & 88.0 \\
CharSwap & 69.4 & 67.5 & 74.1  & 66.4 & 67.2 & 75.4 \\
Distractor Extraction & 48.8 & 47.0 & 44.9 & 89.4 & 60.3 & 89.9 \\
Distractor Generation & 68.9 & 66.6 & 67.2  & 73.3 & 86.5 & 87.8 \\
\midrule
Average & 66.6 & 70.2 & 65.1 & 77.1 & 72.1 & 84.4 \\
\bottomrule
\end{tabular}}
\caption{Results for adversarial training. Each column corresponds to one adversarial training method.}
\label{advTrain}
\end{table*}

\begin{table}[h]
\centering
\small 
\begin{tabular}{ cccc }
\toprule
& \textbf{Orig} & \textbf{AddSent-AT} & \textbf{Mix-AT}
 \\
\midrule
AddSent & 62.3 & 88.0 & 88.0 \\ 
Diagnosis & 95.8 & 19.7 & 11.3 \\ 
\bottomrule
\end{tabular}
\caption{Model performance on AddSent and Diagnosis test set.}
\label{AddSent-AT}
\end{table}

\begin{table}[h]
\centering
\small 
\begin{tabular}{ cccc }
\toprule
& \textbf{Orig} & \textbf{DE} & \textbf{DG}
 \\
\midrule
PQ-remove & 39.8 & 73.8 & 83.7 \\ 
Random & 25.0 & 25.0 & 25.0 \\
\bottomrule
\end{tabular}
\caption{Results for PQ-remove training and testing.}
\label{DE-DG-Analysis}
\end{table}



\begin{table*}[t]
\centering \footnotesize
\begin{tabular}{p{0.005\textwidth}p{0.93\textwidth}}

 \toprule
 \parbox[t]{1mm}{\multirow{2}{*}{\rotatebox[origin=c]{90}{{\textbf{DE}}}}} &
\textbf{Passage:} I personally would be more concerned for the teenager who has to become independent without having any familiarity with working. The biggest reason teens should be allowed to work is that it is a healthy way of earning money. My dad told me, ``Having a job is a good way to save up for things you are going to need or want to do. You get a lot more out of things if you are financially responsible for them.'' Kids need the freedom to choose how to spend their money. The sooner they have an income, the sooner they can learn how to use money wisely. If they are not allowed to work in high school, they may run into trouble in the future. \\ & \textbf{Question:} The author's father advised him to  \_  . \\  & \textbf{Answer:} earn money to get what he wants \\
& \textbf{Predicted:} work  \\ 

\midrule
 \parbox[t]{1mm}{\multirow{2}{*}{\rotatebox[origin=c]{90}{{\textbf{DE}}}}} &
\textbf{Passage:} In a world with limited land, water and other natural resources, the harm from the traditional business model is on the rise. Actually, the past decade has seen more and more forests disappearing and the globe becoming increasingly warm. People now realize that this unhealthy situation must be changed, and that we must be able to develop in sustainable ways. That means growth with low carbon or development of sustainable products. In other words, we should keep the earth healthy while using its supply of natural resources. Today, sustainable development is a proper trend in many countries. According to a recent study, the global market for low-carbon energy will become three times bigger over the next decade. China, for example, has set its mind on leading that market, hoping to seize chances in the new round of the global energy revolution.   \\ & \textbf{Question:} What is the main purpose of the passage? \\  & \textbf{Answer:} To introduce a new business model. \\
& \textbf{Predicted:} sustainable development  \\ 

\midrule
 \parbox[t]{1mm}{\multirow{2}{*}{\rotatebox[origin=c]{90}{{\textbf{DE}}}}} &
\textbf{Passage:} Phyllis Lee of Singapore knew something wasn't right. Her younger son, Alex, then six years old, was getting good grades in his private kindergarten classes. But Lee realized something was wrong when Alex came home one day, crying, with ``zeroes'' on his Chinese spelling test. Lee decided to investigate. According to Lee, the teacher would frequently describe Alex's Chinese writing as 'ghost writings' and made him a laughing stock in class instead of helping him out. Lee, 46, spent the next 12 months teaching Alex and still helps him when necessary. She not only taught him the formation of the Chinese words but also their origin so that he could understand better, often taking more than an hour to read a simple paragraph. Visits to the library and surfing the Internet kept her up to speed on teaching materials. Alex's grades improved and by the end of Year One, he had become one of the top students in Chinese in his class. Alex is keeping an A - plus average in all subjects, and his mother's involvement is a big reason behind it.  \\ & \textbf{Question:} Alex arrived home with tears because \_ . \\  & \textbf{Answer:} he failed in the Chinese exam \\
& \textbf{Predicted:} ``zeroes" on his Chinese spelling test  \\

\midrule
 \parbox[t]{1mm}{\multirow{2}{*}{\rotatebox[origin=c]{90}{{\textbf{DG}}}}} &
\textbf{Passage:} A new survey finds that more than eighty percent of Internet users in the United States search for health information online. The survey found that searching online is one of the leading ways that people look for a second opinion though doctors are still the main source of health information. Forty-four percent of people are actually looking for doctors or other providers when they search for health information online. Another finding of the survey: Two-thirds of Internet users look online for information about a specific disease or medical condition. The Internet has also become an important source of emotional support for people with health problems. Susannah Fox says one in five Internet users has gone online to find other people who have the same condition. It was more popular among people with more serious health issues--one in four people living with chronic diseases. They are so eager to find other people online who share their health concerns. The rise of social networking has made it easier for people with rare diseases to connect with each other and feel less alone.  \\ & \textbf{Question:} By using social networking, patients with rare diseases can  \_  . \\  & \textbf{Answer:} get emotional comfort \\
& \textbf{Predicted:} find other people online  \\ 

\midrule
 \parbox[t]{1mm}{\multirow{2}{*}{\rotatebox[origin=c]{90}{{\textbf{DG}}}}} &
\textbf{Passage:} I was personally discouraged by an adult during my high school. After telling her what university I wanted to attend, she plainly told me I would not get in. I was completely shocked and angry at her statement. The adult may have not intended to hurt me with her words, but it had an after effect. The meeting made me think she had no belief that I could possibly succeed in the future. After that it caused me to try to avoid any future meeting with adults until absolutely necessary. We are all human; therefore, we can all understand that some days are more challenging. But when people allow situations of stress to consume  them, they cannot perform their best. People should pay attention to how their reactions could affect the person they are interacting with.  adults who come off in a rude and aggressive way through communication have an effect on teenagers' mind. Adults whether in schools, or any other institutions should work to tear down walls gently and create a safe space for the person they are serving. These adults should also help and contribute to a person's academic, personal and professional growth. High school is an important time when young people need someone to believe in them. In conclusion, I would like to add that it is not completely up to adults only. Students are responsible for seeking help from adults who are in authority  positions. They are also responsible for the way they approach adults in their academic surroundings. Students can expect to be treated in a respectful way when they express at the beginning. Generally the responsibility lies on both parties. When both sides can communicate in a polite manner then the complete environment of the school has the potential to develop well.  \\ & \textbf{Question:} The passage is mainly about \_  . \\  & \textbf{Answer:} the adults' role in school \\
& \textbf{Predicted:} adults and teenagers  \\ 

\bottomrule
\end{tabular}
\caption{Examples of DE and DG that successfully fooled RoBERTa model.}
\label{tab:DE_DG_examples}
\end{table*}

\section{Adversarial Training}
\label{sec:at}
Adversarial training (AT) is a common method to improve the robustness of NLP models where adversarial examples are added into the training data. In this section, we first examine how different AT strategies improve the model performance on various adversarial attacks. Next, we analyse the reasons behind the performance improvement and highlight important flaws of using adversarial training as defense. 

\subsection{Adversarial Training Experiments}
Previous works have used adversarial training to improve model robustness under specific attacks~\cite{AdvSquad, AddSentDiverse}. In this section, we first explore this targeted adversarial training setting where we randomly sample 25\% of training data, apply the attack on the sampled training data, and add the perturbed data into the original training set as augmentation.
Furthermore, we explore a mix adversarial training setting, where we add all the four types of adversarial data into the original training set together. In this setting, each adversarial attack is applied to 25\% of the training data, so the resultant augmented training data has double size of the original training set.

We use RoBERTa for all the adversarial training experiments with the same hyperparameters.
The experiment results are shown in Table~\ref{advTrain}, where the first column is the model's performance on the original RACE test set, the middle four columns are results of targeted adversarial training where each of the individual adversarial attack data is added in training. The last column shows results for the mix adversarial training where all the four types of AT data are added during training. We outline a few interesting findings from our experiments:
\begin{itemize}\setlength{\itemsep}{0.01pt}
    \item Adversarial training on one type of adversarial data only improves model's performance on the attack that the model is adversarially trained on, while generally decreasing its performance on other test sets.
    \item Adversarial training on Distractor Extraction improves the model's performance on AddSent as well, possibly because that it makes the model rely less on simple text matching.
    Adversarial training on Distractor Generation significantly improves the model's performance on the DE test set, while adversarial training on DE also moderately improves the model performance on the DG test set, possibly due to the similar distractor replacement format of the two attack methods.
    \item Mix Adversarial training significantly improves the model's performance on all the adversarial test sets with slight drop on the original test set.
\end{itemize}

\subsection{Understanding the effects of Adversarial Training}
While it appears that adversarial training has significantly improved the model's performance under adversarial attacks, it is unclear whether it has really made the model more robust. Moreover, we find that on AddSent, DE, and DG, the adversarially augmented model achieves higher performance on the adversarial test set than the original test set, which may suggest overfitting.
Hence, we analyse what is learned by the model during AT.

\noindent 
\subsubsection*{$\bullet$~~ AddSent Adversarial Training}
We hypothesize that there exists spurious patterns in the AT data which are exploited by the model. 
To test this hypothesis, we construct a diagnosis test set by inserting the concatenation of the question and the correct answer into the passage, which does not change the labels and adds additional lexical matching for the correct answer. In Table~\ref{AddSent-AT}, three models: RoBERTa finetuned on the original training set (Orig), under AddSent adversarial training (AddSent-AT), and under mix adversarial training (Mix-AT) are evaluated on the AddSent test set as well as our diagnosis test set. Comparing the first line with the second line, we find that 
RoBERTa finetuned on the original training set achieves much higher accuracy on the diagnosis test test, indicating that the insertion of correct answers makes it easier for the model to identify the correct answer. However, AddSent-AT and Mix-AT achieve miserable performance on the diagnosis test set, indicating such adversarially trained models are exploiting spurious patterns to score well and the model will fail when such patterns do not apply in the new test set. Hence it is likely that the high performance after AT is an overestimate, and better measures should be taken as defense to ensure improvement in the model's robustness.

\noindent 
\subsubsection*{$\bullet$~~ DE and DG Adversarial Training}
To examine the exploiting of spurious statistical cues in the extracted and generated distractors, we employ partial data training~\cite{partialMRC, Si2019WhatDB} where we train and test RoBERTa on RACE without the passages and questions (i.e., we only input the options to the model). Under this setting, it is impossible for humans to correctly answer the questions since only the options are provided. As shown in Table~\ref{DE-DG-Analysis}, we train RoBERTa on: the original training set, DE adversarial training data, and DG adversarial training data, and test on their respective test sets. Under such option-only (PQ-remove) setting, RoBERTa achieves performance that is much higher than random on the original, DE, and DG test sets. This indicates that all the three datasets (original, DE and DG) have spurious statistical cues that are exploited to achieve high performance in the option-only setting. Moreover, models are exploiting such cues in DE and DG datasets to a much larger extent compared to on the original dataset. Thus, adding these examples into training may lead to an overestimate of the model's performance as the model can easily exploit these spurious patterns to increase the performance. 

In summary, we find that the adversarially trained models may not be more robust, and the high scores achieved by adversarial training may have overestimated the models' robustness. We believe that more reliable defense methods should be developed to improve the model's robustness without exploiting spurious statistical cues as a shortcut. We leave such exploration to future work.

\section{Additional Examples}

We present more examples of our DE and DG attacks in Table~\ref{tab:DE_DG_examples}.

